# A Two-Phase Genetic Algorithm for Image Registration

Sarit Chicotay  
Bar-Ilan University  
Ramat-Gan, Israel  
saritc@gmail.com

Eli (Omid) David[1]  
Bar-Ilan University  
Ramat-Gan, Israel  
mail@elidavid.com

Nathan S. Netanyahu[2]  
Bar-Ilan University  
Ramat-Gan, Israel  
nathan@cs.biu.ac.il

## ABSTRACT

*Image Registration* (IR) is the process of aligning two (or more) images of the same scene taken at different times, different viewpoints and/or by different sensors. It is an important, crucial step in various image analysis tasks where multiple data sources are integrated/fused, in order to extract high-level information.

Registration methods usually assume a relevant transformation model for a given problem domain. The goal is to search for the "optimal" instance of the transformation model assumed with respect to a similarity measure in question.

In this paper we present a novel *genetic algorithm* (GA)-based approach for IR. Since GA performs effective search in various optimization problems, it could prove useful also for IR. Indeed, various GAs have been proposed for IR. However, most of them assume certain constraints, which simplify the transformation model, restrict the search space or make additional preprocessing requirements. In contrast, we present a generalized GA-based solution for an almost fully affine transformation model, which achieves competitive results without such limitations using a two-phase method and a *multi-objective optimization* (MOO) approach.

We present good results for multiple dataset and demonstrate the robustness of our method in the presence of noisy data.

## 1   INTRODUCTION

Image registration (IR) is an important, significant component in many practical problem domains. Due to the enormous diversity of IR applications and methodologies, automatic IR remains a challenge to this day. A broad range of registration techniques has been developed for various types of datasets and problem domains [1], where typically, domain-specific knowledge is taken into account and certain a priori assumptions are made to simplify the model in question.

An affine transformation is one of the most commonly used models. Since the search space is too large for a feasible exhaustive search through the entire parameter space, the major challenge is to avoid getting stuck at a local optimum when there are multiple extrema in the similarity metric search space.

In order to overcome this problem, we present in this paper a novel two-phase *genetic algorithm* (GA)-based approach for IR. We devise a GA-based framework coupled with image processing techniques to search efficiently for an optimal transformation with respect to a given similarity measure. Due to our two-phase strategy and a unique simultaneous optimization of two similarity measures based on a *multi-objective optimization* (MOO) strategy, we obtain good results over a relatively large search space assuming an almost fully affine transformation model.

## 2   TWO-PHASE GA-BASED IR

This section describes briefly our two-phase GA-based approach to optimize the devised similarity measures by utilizing common IR tools. For a detailed presentation of this work see [2].

Our IR scheme searches for a transformation that generates a maximal match in the overlap between the reference image and the transformed sensed image, thus, the GA chromosome is defined by six genes reflecting the effects represented by an affine transformation; translation along the $x$-and-$y$ axis, rotation, scale factor, and shear along the $x$-and-$y$ axis.

Two similarity measures are used in this work: (1) *Euclidean distance* measure, which is applied to geometric feature points extracted from both images, and (2) *normalized cross correlation* (NCC) [3], which is an intensity-based measure.

The Euclidean distance measure computes the similarity between two feature point sets, $P$ and $Q$, extracted from the reference and sensed image. We first tested our scheme using manually selected features and showed that without assuming correspondences our algorithm gives good registration results. We then applied the measure to wavelet features obtained in a fully-automatic mode from Simoncelli's steerable filters [4] based on a wavelet transform. The Euclidean distance measure is calculated for the two extracted feature sets as follows:

First, the feature points extracted from the sensed image are warped according to the transformation assumed. For each warped point $\vec{p}$ we determine its corresponding point $\vec{q}$ among the unassigned reference feature points $Q' \subseteq Q$, by finding its nearest neighbor with respect to Euclidean distance $d(\vec{p}, \vec{q})$, i.e.,

---

[1] www.elidavid.com  
[2] Nathan Netanyahu is also with the Center for Automation Research, University of Maryland, College Park, MD 20742.

$$corr(p) = \left\{\vec{q}\epsilon Q' \,\Big|\, \min_{\vec{q}\epsilon Q'} d(p,q)\right\}$$

Finally, the similarity value is the value of the median Euclidean distance among the correspondences found, i.e.,

$$d(P,Q) = med_{\vec{p}\epsilon P}\left(d(\vec{p}, corr(\vec{p}))\right)$$

The second measure used is the normalized cross correlation (NCC), which has been commonly applied to evaluate the degree of similarity between two images. For image $A$ and the warped image $B$ it is defined as:

$$NCC(A,B) = \frac{\sum_{x,y}(A(x,y) - \bar{A})(B(x,y) - \bar{B})}{\sqrt{\sum_{x,y}(A(x,y) - \bar{A})^2 \sum_{x,y}(B(x,y) - \bar{B})^2}}$$

where $\bar{A}$ and $\bar{B}$ are the average gray levels of images $A$ and $B$.

Having performed several tests using each of these measures, independently, as the fitness function, we noticed that the GA fails to obtain consistent results with a single measure. In an attempt to obtain a more robust IR scheme, we combined, therefore, the two measures as part of our two-phase strategy.

## 2.1 Phase 1: Coarse Estimation

The goal of the first phase is to obtain an initial coarse estimate. This is achieved using the Euclidean distance measure which is expected to yield consistent candidate solutions that are "relatively" close to the optimal solution.

The first phase completes when there is no "significant" update for a predefined number of changes or when converging to some predefined minimal distance measure.

## 2.2 Phase 2: Multi-Objective Optimization

The second phase starts with the population at the end of the first phase. The Euclidean distance measure is combined with the NCC measure which makes use of the full image data.

Ideally, we would like to optimize simultaneously the two objective functions, however, in practice, they may not be optimized simultaneously. Thus, we use a multi-objective optimization approach that gives a partial ordering of solutions based on *Pareto dominance*.

The second phase completes when there is no "significant" update. We select among the *pareto-optimal set* the individual with the best NCC value as the suggested solution.

## 3 EMPIRICAL RESULTS

We tested our algorithm on a few dozens of synthetic and real image datasets, including various satellite and outdoor scenes, in both a semi-automatic and a fully-automatic mode.

The correctness of the final transformation is evaluated by the root mean square error (RMSE) for manually selected points. We consider RMSE value < 1.5 pixels as a good registration.

The semi-automatic mode yields good results in all of the cases considered. The tests in a fully-automatic mode achieved successful registration in RMSE terms in about 75% of the test cases. Some of the failed cases can be recovered, though, if additional measures/constraints are applied to the transformation's parameters, e.g., using *mutual information* (MI) instead of NCC (affected by contrast relationships). Table 1 and Figure 1 present several results in a fully-automatic mode.

We compared also our results on multiple datasets to other methods assuming a simpler transformation model and performed additional tests on real images from INRIA database [5] that underwent affine transformations. See [2] for full details.

## 4 CONCLUSIONS

In this paper we presented a novel two-phase GA-based image registration algorithm, whose main advantage over existing evolutionary IR techniques is that it provides a robust and automatic solution for a (quasi) fully affine transformation which is one of the most commonly used models for image registration. We used the Euclidean distance measure and the NCC measure as part of a two-phase strategy, supported by a novel MOO design, which is used to optimize the two similarity measures simultaneously.

We have tested extensively the proposed scheme, and demonstrated its robustness to noisy data and consistency on multiple datasets over successive runs.

Further research should be done to achieve a robust, fully-automatic registration in more challenging scenarios.

**Table 1: Fully-automatic registration results of the images in Figure 1 (RMSE in pixels).**

| Image | Avg. RMSE | σ RMSE |
|---|---|---|
| Boat | 1.37 | 0.2 |
| House | 1.34 | 0.24 |

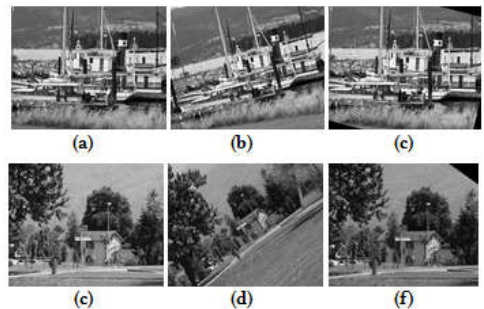

**Figure 1: (a), (d) Reference, (b), (e) sensed, and (c), (f) registered images of "boat" and "house" pairs from [5].**